\documentclass[letterpaper, 10 pt, conference]{ieeeconf}  % Comment this line out if you need a4paper

\IEEEoverridecommandlockouts             % This command is only needed if 
\usepackage{graphicx} % for pdf, bitmapped graphics files
\title{\LARGE \bf
A Study of Perceived Safety for Soft Robotics in Caregiving Tasks
}

\author{Cosima du Pasquier, Jennifer Grannen, Chuer Pan, Serin L. Huber, Aliyah Smith,\\ Monroe Kennedy, Shuran Song, Dorsa Sadigh, and Allison M. Okamura% <-this % stops a space
\thanks{This work was supported in part by the Stanford Robotics Center (SRC).}% <-this % stops a space
\thanks{The authors are with the Department of Mechanical Engineering and the Department of Computer Science, Stanford University, Stanford, CA, 94305. Email:
\{cosimad, jgrannen, chuerpan, huberser, aliyah1, monroek, shuran, dorsa, aokamura\}@stanford.edu}%
}

\begin{document}

\maketitle
\thispagestyle{empty}
\pagestyle{empty}

%%%%%%%%%%%%%%%%%%%%%%%%%%%%%%%%%%%%%%%%%%%%%%%%%%%%%%%%%%%%%%%%%%%%%%%%%%%%%%%%
\begin{abstract}

In this project, we focus on human-robot interaction in caregiving scenarios like bathing, where physical contact is inevitable and necessary for proper task execution because force must be applied to the skin. Using finite element analysis, we designed a 3D-printed gripper combining positive and negative pressure for secure yet compliant handling. Preliminary tests showed it exerted a lower, more uniform pressure profile than a standard rigid gripper. In a user study, participants' trust in robots significantly increased after they experienced a brief bathing demonstration performed by a robotic arm equipped with the soft gripper. These results suggest that soft robotics can enhance perceived safety and acceptance in intimate caregiving scenarios.

\end{abstract}

%%%%%%%%%%%%%%%%%%%%%%%%%%%%%%%%%%%%%%%%%%%%%%%%%%%%%%%%%%%%%%%%%%%%%%%%%%%%%%%%
\section{Introduction}
\label{Section I}
Despite two decades of innovation, soft robotics still struggles to find practical, high-impact applications beyond academic research~\cite{Roh2024}. The few real-world uses consist mainly of handling delicate objects. Commonly cited challenges include predicting deformation and fatigue in soft materials, establishing robust and scalable fabrication methods, and integrating sensing for precise control~\cite{Blanco2024,Hegde2023}. Beyond manipulation tasks, there is no definitive example showcasing how soft robots’ unique properties can transform an application at a large scale. \par

Healthcare is frequently mentioned as a promising arena for soft robots, given their potential to provide safer interactions with patients; because soft materials have stiffness levels similar to human skin, they reduce mechanical mismatching~\cite{Cianchetti2016,Wang2024Pioneering,Roh2024}. However, the claim that soft robots inherently improve safety in human-robot interaction remains largely qualitative. It is also unclear whether the wider public is prepared to accept the notion that soft robotics offer a greater degree of safety. \par

As healthcare systems face rising patient demands, aging populations, and staff shortages, robotics and automation become crucial for maintaining patient care and preserving autonomy and dignity. However, it is essential to evaluate whether the general population will trust and accept soft robots in intimate, personal settings before advancing these technologies.\par

Two recent studies investigated human perception in human soft-robot interaction (HSRI). The first study examined whether soft robots felt more natural than rigid tentacles~\cite{Jorgensen2022}. Although this was not confirmed, participants interacted more freely with the soft robots, suggesting a sense of safety. In the second study~\cite{Wang2024}, participants interacted with a soft robotic hand performing human-like actions on their forearms and completed a safety questionnaire. Both studies provided safety insights but remained exploratory, failing to conclusively show that soft robots enhance perceived safety in interactions. \par

In this study, we examined whether a soft robot can improve perceived safety in caregiving tasks like bathing. We designed and tested a 3D-printed soft gripper for holding bathing implements such as sponges and towels. Participants performed a bathing task with a robotic arm equipped with the gripper and completed pre- and post-interaction surveys to evaluate their familiarity, comfort, and trust in robots.

\section{Gripper Design, Fabrication, and Testing}
\label{Section II}
We used finite element analysis (FEA) to simulate and design a soft pneumatic gripper. Each finger included three linear actuators to enable dexterous motion and a secure grip. The three-finger design includes actuators, end-effectors, and connecting elements that also function as air channels. \par

We used stereolithography (SLA, Formlabs, Form 3) to iterate on the shape of each component. We tested three soft 3D-printing techniques for our components: SLA, embedded printing (Raplid Liquid Printing), and digital light processing (DLP, Carbon, M2). DLP was ultimately chosen for its superior geometric accuracy, durability, and ease of assembly. Each component was printed separately using EPU 40, cured for 30 minutes at 175$^\circ$, assembled with a custom jig, and cured for the remaining 11.5 hours. \par

The gripper was mounted to a 7-DoF Kinova Gen 3 robotic arm and was actuated using a custom positive and negative pressure controller. We used an Intel RealSense RGBD camera as an third-person overhead camera, with arm skeleton detection and segmentation from OpenCV to detect arm location and position on a table, which determined the end effector trajectory for the gripper to follow. \par

\begin{figure}
    \centering
    \includegraphics[width=0.8\linewidth]{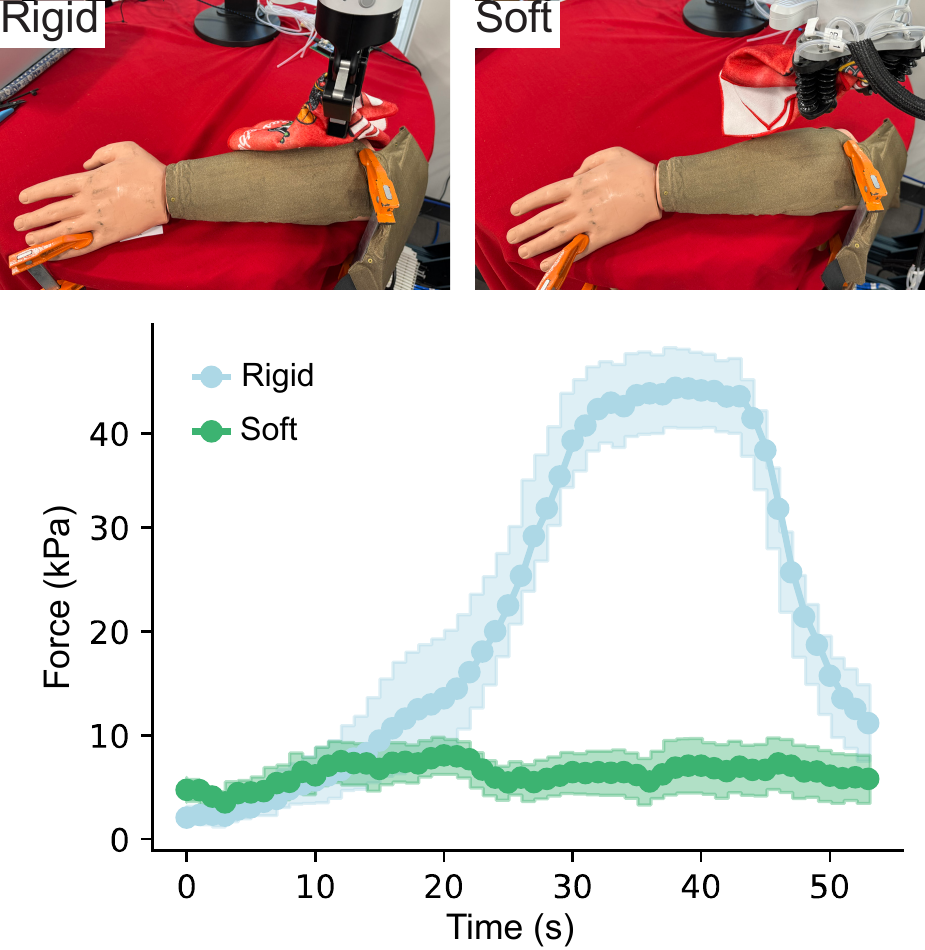}
    \caption{(Top) Test setup for preliminary force testing using a rigid (left) and soft (right) gripper to brush a towel on a mannequin arm; (Bottom) average force profile of n=3 tests for both grippers.}
    \label{fig:preliminary_testing}
\end{figure}

To confirm that the soft gripper applied less force than a rigid gripper, we conducted preliminary tests on a mannequin arm using a sensor array (Pressure Profile Systems). Both a standard Robotiq gripper and our soft gripper were driven along the arm. As shown in Fig.~\ref{fig:preliminary_testing}, the soft gripper exerted a lower, more uniform pressure profile on the arm.

\section{User Testing}
\label{Section III}
We conducted a user study with 106 participants ages 18 to 89 to evaluate whether interacting with a soft robotic gripper in a caregiving scenario would increase trust in robots. Participants were recruited during the Stanford Robotics Center launch in November 2024. The protocol was approved by the Institutional Review Board.\par

The test consisted of four phases. First, observers received an overview of the demonstration and were invited to participate. Those who agreed completed a consent protocol and filled out a pre-experiment survey. Next, participants sat at a table and positioned their arm on marked locations. Once the arm position was confirmed, the robotic arm moved to the participant’s elbow, swept a towel held by the gripper along their arm three times, and returned to its starting position. Finally, participants completed a post-experiment survey.\par

The pre-experiment survey included demographic questions, assessed participants' initial familiarity and level of trust in robots using a 7-point Likert scale. The post-experiment survey repeated the same questions concerning trust in robots, and had open ended questions concerning participants' experience. 

\section{Results and Discussion}
\label{Section IV}
\begin{figure}
    \centering
    \includegraphics[width=\linewidth]{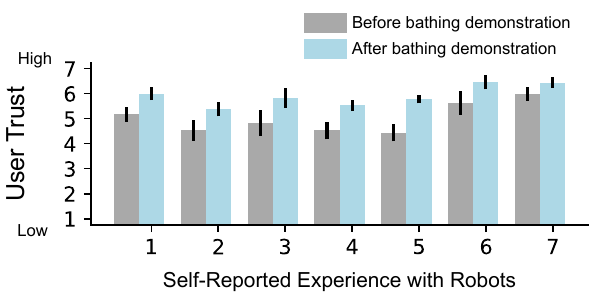}
    \caption{Self-reported user trust in robots against experience with robots before and after the bathing demonstration on a 7-point Likert scale. }
    \label{fig:trust}
\end{figure}

When comparing participants’ overall trust ratings before and after the experiment, the proportion giving the highest rating (7) increased from 21.3\% to 39.6\%, while the cumulative proportion assigning the lowest ratings (1–3) declined from 14.8\% to 2.8\%. Figure~\ref{fig:trust} shows the average trust rating by self-reported experience level, ranging from no experience (1) to expert (7). Although trust rose across all categories, the increase was smallest among experts, who had already indicated high initial trust. Interestingly, participants with the least (1) and most (7) experience reported the highest baseline trust, whereas those in the middle were more cautious; however, this difference largely disappeared after the experiment.\par

This experiment demonstrates that interacting with a soft robot in a caregiving scenario increases trust in robots among individuals with varying levels of expertise. However, the study was limited in scope, given it had to be completed in under five minutes. We are currently refining the control system—particularly its sensing capabilities—to detect contact more effectively. We also plan to conduct a second, more extensive user study, which will more clearly evaluate the efficiency of a soft robotic gripper in caregiving, such as the inherent safety provided by material compliance rather than relying solely on software-based safeguards.

%%%%%%%%%%%%%%%%%%%%%%%%%%%%%%%%%%%%%%%%%%%%%%%%%%%%%%%%%%%%%%%%%%%%%%%%%%%%%%%%

%%%%%%%%%%%%%%%%%%%%%%%%%%%%%%%%%%%%%%%%%%%%%%%%%%%%%%%%%%%%%%%%%%%%%%%%%%%%%%%%



\section*{Acknowledgments}

The authors thank Roger Wise at the Additive Manufacturing and Prototyping Facility at Stanford University for his support and advice in printing the soft gripper.

%%%%%%%%%%%%%%%%%%%%%%%%%%%%%%%%%%%%%%%%%%%%%%%%%%%%%%%%%%%%%%%%%%%%%%%%%%%%%%%%

%\begin{thebibliography}{99}
% \bibitem{c1} {(bib is having issues)}
\bibliography{bib}

\begin{thebibliography}{1}
\providecommand{\url}[1]{#1}
\csname url@samestyle\endcsname
\providecommand{\newblock}{\relax}
\providecommand{\bibinfo}[2]{#2}
\providecommand{\BIBentrySTDinterwordspacing}{\spaceskip=0pt\relax}
\providecommand{\BIBentryALTinterwordstretchfactor}{4}
\providecommand{\BIBentryALTinterwordspacing}{\spaceskip=\fontdimen2\font plus
\BIBentryALTinterwordstretchfactor\fontdimen3\font minus \fontdimen4\font\relax}
\providecommand{\BIBforeignlanguage}[2]{{%
\expandafter\ifx\csname l@#1\endcsname\relax
\typeout{** WARNING: IEEEtran.bst: No hyphenation pattern has been}%
\typeout{** loaded for the language `#1'. Using the pattern for}%
\typeout{** the default language instead.}%
\else
\language=\csname l@#1\endcsname
\fi
#2}}
\providecommand{\BIBdecl}{\relax}
\BIBdecl

\bibitem{Roh2024}
Y.~Roh, Y.~Lee, D.~Lim, D.~Gong, S.~Hwang, M.~Kang, D.~Kim, J.~Cho, G.~Kwon, D.~Kang, S.~Han, and S.~H. Ko, ``Nature's blueprint in bioinspired materials for robotics,'' \emph{Advanced Functional Materials}, vol.~34, p. 2306079, 2024.

\bibitem{Blanco2024}
K.~Blanco, E.~Navas, L.~Emmi, and R.~Fernandez, ``Manufacturing of 3d printed soft grippers: A review,'' \emph{IEEE Access}, vol.~12, pp. 30\,434--30\,451, 2024.

\bibitem{Hegde2023}
C.~Hegde, J.~Su, J.~M.~R. Tan, K.~He, X.~Chen, and S.~Magdassi, ``Sensing in soft robotics,'' \emph{ACS Nano}, vol.~17, pp. 15\,277--15\,307, 2023.

\bibitem{Cianchetti2016}
M.~Cianchetti and C.~Laschi, ``Pleasant to the touch: By emulating nature, scientists hope to find innovative new uses for soft robotics in health-care technology,'' \emph{IEEE Pulse}, vol.~7, pp. 34--37, 5 2016.

\bibitem{Wang2024Pioneering}
Y.~Wang, Z.~Xie, H.~Huang, and X.~Liang, ``Pioneering healthcare with soft robotic devices: A review,'' \emph{Smart Medicine}, vol.~3, p. e20230045, 2024.

\bibitem{Jorgensen2022}
J.~Jørgensen, K.~B. Bojesen, and E.~Jochum, ``Is a soft robot more “natural”? exploring the perception of soft robotics in human–robot interaction,'' \emph{International Journal of Social Robotics}, vol.~14, pp. 95--113, 1 2022.

\bibitem{Wang2024}
Y.~Wang, G.~Wang, W.~Ge, J.~Duan, Z.~Chen, and L.~Wen, ``Perceived safety assessment of interactive motions in human–soft robot interaction,'' \emph{Biomimetics}, vol.~9, p.~58, 2024.

\end{thebibliography}
%\end{thebibliography}
\end{document}